\documentclass[sigconf]{acmart}
\AtBeginDocument{%
  }

\copyrightyear{2025}
\acmYear{2025}
\setcopyright{acmlicensed}
\acmConference[UAVM '25] {Proceedings of the 3rd International Workshop on UAVs in Multimedia: Capturing the World from a New Perspective}{October 27--31, 2025}{Dublin, Ireland.}
\acmBooktitle{Proceedings of the 3rd International Workshop on UAVs in Multimedia: Capturing the World from a New Perspective (UAVM '25), October 27--31, 2025, Dublin, Ireland}
\acmISBN{979-8-4007-1839-7/2025/10}
\acmDOI{https://doi.org/10.1145/3728482.3757392.}

\usepackage{pifont}
\usepackage{bbding}
\usepackage{fontawesome}
\usepackage[marginal]{footmisc}

\begin{document}

\title{SkyLink: Unifying Street-Satellite Geo-Localization via UAV-Mediated 3D Scene Alignment}

\author{  
Hongyang Zhang}
\authornote{Both authors contributed equally to this research.}
\affiliation{%
  \institution{Key Laboratory of Multimedia Trusted Perception and Efficient Computing, Ministry of Education of China, Xiamen University}
  \city{Xiamen}
  \state{Fujian}
  \country{China}
  \postcode{361102}}
\email{hyzhang@stu.xmu.edu.cn}

\author{  
Yinhao Liu}
\authornotemark[1]
\affiliation{%
  \institution{Key Laboratory of Multimedia Trusted Perception and Efficient Computing, Ministry of Education of China, Xiamen University}
  \city{Xiamen}
  \state{Fujian}
  \country{China}
  \postcode{361102}}
\email{liuyinhao28@stu.xmu.edu.cn}

\author{  
Zhenyu Kuang}
\authornote{Corresponding author.}
\affiliation{%
  \institution{School of Electronic and Information Engineering, Foshan University}
  \city{Foshan}
  \state{Guangdong}
  \country{China}
  \postcode{528200}}
\email{kuangzhenyu@fosu.edu.cn}

\renewcommand{\shortauthors}{Hongyang Zhang, Yinhao Liu, and Zhenyu Kuang}

\begin{abstract}
Cross-view geo-localization aims at establishing location correspondences between different viewpoints. Existing approaches typically learn cross-view correlations through direct feature similarity matching, often overlooking semantic degradation caused by extreme viewpoint disparities. To address this unique problem, we focus on robust feature retrieval under viewpoint variation and propose the novel SkyLink method. We firstly utilize the Google Retrieval Enhancement Module to perform data enhancement on street images, which mitigates the occlusion of the key target due to restricted street viewpoints. The Patch-Aware Feature Aggregation module is further adopted to emphasize multiple local feature aggregations to ensure the consistent feature extraction across viewpoints. Meanwhile, we integrate the 3D scene information constructed from multi-scale UAV images as a bridge between street and satellite viewpoints, and perform feature alignment through self-supervised and cross-view contrastive learning. Experimental results demonstrate robustness and generalization across diverse urban scenarios, which achieve 25.75$\%$ Recall@1 accuracy on University-1652 in the UAVM2025 Challenge. Code will be released at \url{https://github.com/HRT00/CVGL-3D}.
\end{abstract}

%
%




\maketitle

\section{Introduction}
\begin{figure}[t]
\begin{center}
     \includegraphics[width=0.95\linewidth]{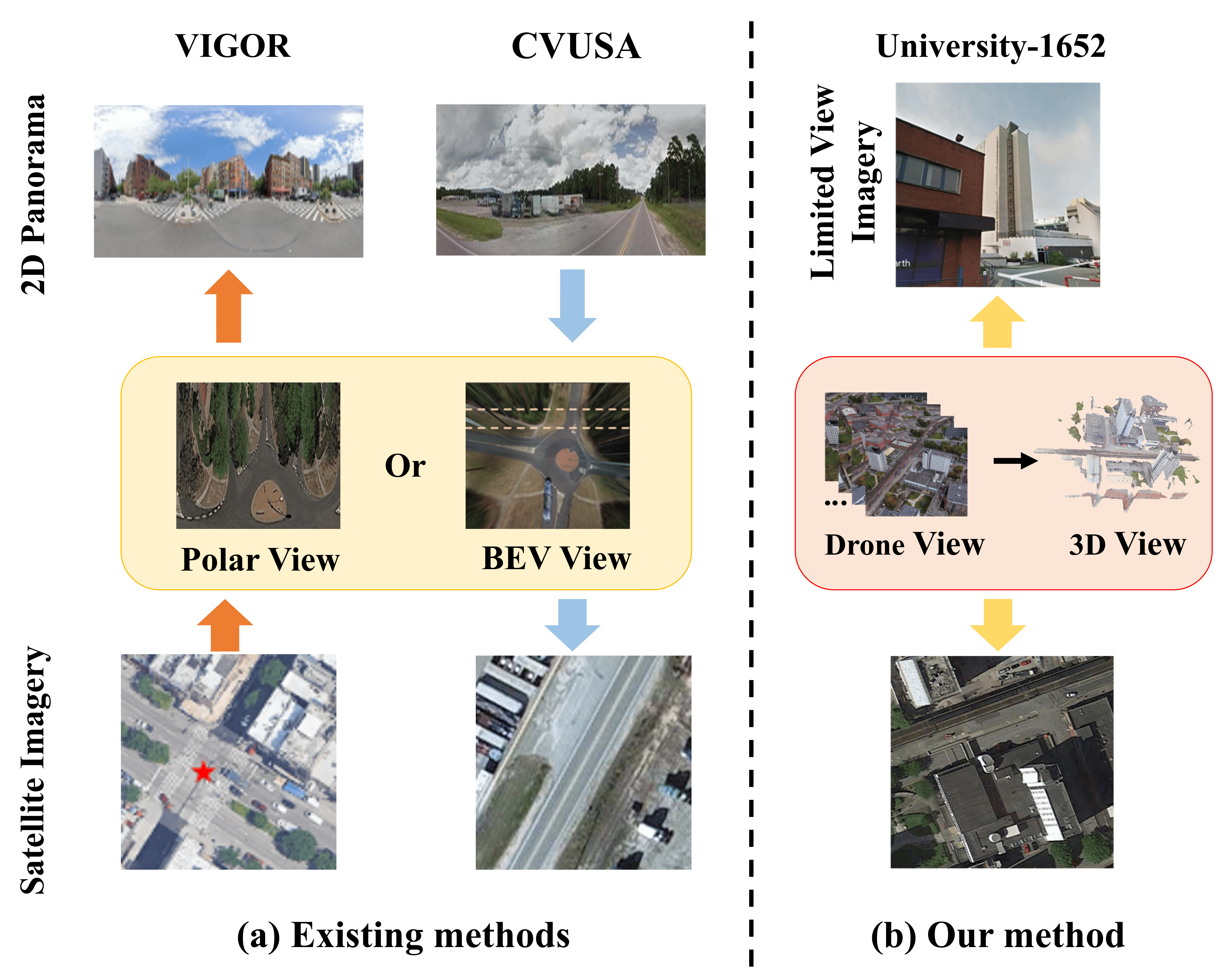}
\end{center}
\vspace{-.1in}
      \caption{(a) Existing methods often adopt polar or BEV transformations for street-satellite matching, but these introduce geometric distortion and lack robustness to viewpoint changes. (b) We utilize 3D point cloud representations from multi-view UAV imagery to bridge the perspective gap. These 3D structures retain fine-grained spatial geometry, improving semantic alignment across views. By introducing a shared 3D space between street and satellite imagery, our method enhances generalization under diverse urban scenarios.
      }\label{fig:visual}
\vspace{-.2in}
\end{figure}

Cross-view geo-localization (CVGL)~\cite{r25, r26, r22} aims to determine the geographic location of street images by matching them with corresponding satellite/drone imagery (or vice versa), which faces significant challenges due to the substantial domain gap between them~\cite{r17}. The drastic differences in viewpoint, scale, and imaging conditions create severe appearance discrepancies that hinder semantic alignment. As shown in Figure~\ref{fig:visual}, existing benchmarks such as University-1652~\cite{r11} offer street-view images with limited fields of view, which often lead to occlusion issues in street-to-satellite matching~\cite{r26}. While more recent datasets like VIGOR~\cite{r13} and CVUSA~\cite{r14} have advanced the field by introducing panoramic views to synthesize satellite-like images and mitigate the cross-view gap, they still suffer from notable limitations, including insufficient viewpoint diversity, alignment inaccuracies, and constrained viewing angles. These drawbacks hinder the learning of robust and view-invariant representations.


\begin{figure*}[t]
\begin{center}
    \includegraphics[width=1\linewidth]{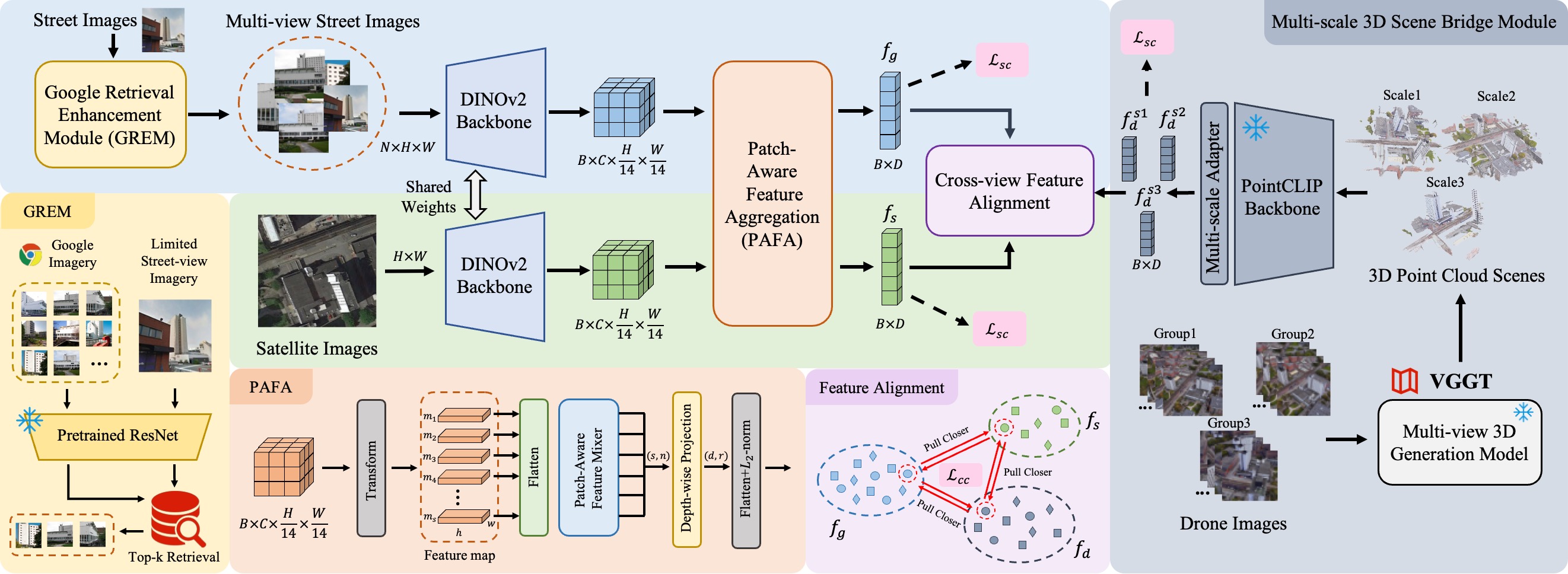}
\end{center}
     \caption{Overview of our proposed framework. The GREM module is first adopted to augment the limited street view data. We then utilize a Siamese network to extract features from the enhanced street and satellite imagery, with local feature aggregation across viewpoints performed by the PAFA module. Additionally, we reconstruct multiple drone images into multi-scale 3D pointcloud scenes, leveraging their embedded representations to bridge the street-view and satellite-view domains. Cross-view features are jointly optimized using self-supervised and cross-view constractive learning.
     }\label{fig:overview}
\end{figure*}

Despite these challenges, key structural elements—including road networks, building contours, and vegetation patterns—exhibit consistent cross-view visibility, offering sparse but semantically rich correspondences. Recent works \cite{r1, r23} demonstrate that street-view panoramas can be effectively transformed into Bird's Eye View (BEV) imagery through azimuth-aware projection with ground-plane constraints, yielding more natural spatial alignment than conventional polar methods \cite{r15} while better preserving local geometries. While both BEV and polar projections suffer from inherent distortion and perspective limitations, 3D point clouds offer an effective solution by capturing detailed spatial layouts and enabling robust multi-view fusion. This capability facilitates a unified representation space bridging street and satellite imagery. Our approach leverages UAV-based multi-scale 3D reconstruction to address critical semantic and geometric disparities between street and satellite views, ultimately enabling more precise cross-view geo-localization.

Based on the aforementioned issues, we propose the SkyLink method. Specifically, considering the limited viewpoints of street-view data, we first enhance the imagery using Google Imagery. We then employ a weight-sharing Siamese network to extract features from both street and satellite images, integrating spatial elements through a patch-aware feature aggregation module. To bridge the cross-view difference, we further reconstruct multi-scale point cloud scenes from drone imagery, whose hierarchical representations help align street and satellite perspectives. A cross-view feature alignment module subsequently constrains the representations from the three viewpoints. This approach leverages rich UAV viewpoint information to build a bridge connecting street and satellite views, ensuring semantic consistency in the high-level feature space. The supreme performance on the University-1652 dataset further demonstrates the effectiveness of our approach.

\section{Proposed Method} \label{method}

In this section, we present SkyLink method to ``link'' street and satellite viewpoints by leveraging multi-scale drone images from the ``sky''. The overall architecture is shown in Figure~\ref{fig:overview}. Formulating CVGL as a metric learning task~\cite{r18, r19, r20}, we optimize feature extractor $F(\cdot)$ by maximizing the similarity between query and gallery embeddings. For the UAVM2025 Challenge, we leverage UAV images at different scales to generate a 3D pointcloud scene through reconstruction, using its representation $f_d$ as an intermediate bridge between street and satellite views.

\subsection{Google Retrieval Enhancement Module}
Street-view images in the competition suffer from limited perspective diversity, all captured from a single ground-level viewpoint. This constraint introduces significant challenges when aligning street-view images with top-down satellite imagery in cross-view geo-localization tasks. To mitigate this issue, we design a Google Retrieval Enhancement Module (GREM) to enrich the visual diversity of street-view samples. Specifically, we first employ a frozen pretrained ResNet-50 model to extract global visual features from all available street-view candidates. Then, for each original street-view image, we compute the cosine similarity between its feature and those of the candidate set, and select the top 50$\%$ most similar images to serve as additional multi-view samples. These retrieved views, often from nearby but slightly different angles, are incorporated into the training process to improve viewpoint robustness.

\subsection{Feature Extraction}

To fully capture the detailed representatioinal information in street images and satellite images, we follow the work of Zheng et al.~\cite{r11} and employ a weight-sharing Siamese architecture for feature extraction. The network is jointly optimized by using information from different viewpoints to ensure the learning of similar target representations such as buildings and roads. 

\subsubsection{Backbone} 

We employ DINOv2~\cite{r5} as our backbone, a vision foundatioin model pretrained on large-scale visual data. Its strong generalization capability enables effective transfer to CVGL tasks while handling challenges from diverse perspectives and complex scenes. The input contains enhanced street images and satellite images with size of $H\times W$. They are independently embedded by a patch embed module using 2D convolutional layers with kernel size of $14 \times 14$. Theses patches are then processed through the ViT blocks of DINOv2, with the number of blocks varing by model size. The output features are $\frac{H}{14}\times \frac{W}{14}$ with channel $C$ of 1024. 

\subsubsection{Patch-Aware Feature Aggregation}

To further enhance cross-view feature interaction and improve model's viewpoint robustness, we propose the patch-aware feature aggregation (PAFA) module. Through a cascade of feature mixing, the relationships between patch elements in the feature maps of different viewpoints are integrated. Specifically, we transform the feature matrix $F$ of size $\frac{H}{14}\times \frac{W}{14}$ into $s$ feature patches $\{{p^i}\}_{i=1}^s$ of dimension $h\times w$. Each feature patch is flattened and then fed into the patch-aware feature mixer module. The module utilizes a stack of isotropic patches composed of multilayer perceptrons, which incorporates spatial global relations into each patch $p^i\in F$ iteratively:
\begin{equation}
    \hat{p}^i \leftarrow W_2(\sigma(W_1p^i))+p^i,
\end{equation}
where $W_1$ and $W_2$ are weights in MLP, and $\sigma(\cdot)$ is the ReLU activation fuction. The feature patches are the aggregated to obtain a feature map $Z\in \mathbb{R}^{s\times n}$, where $n=h\times w$. We then transform the channel dimensions and rows through the depth-wise projection layers to obtain $Z'\in\mathbb{R}^{d\times r}$, which is finally flattened to a $D$-dimensional feature vector $f$.

\subsection{Multi-scale 3D Scene Bridge Module}

Following~\cite{r10}, we observe that high-altitude drone captures approximate satellite imagery, while low-altitude shots retain street-level details. This unique characteristic motivates us to effectively bridge the cross-view semantic gap through UAV-mediated feature learning.

The drone images are divided into three groups, each containing 18 consecutive images categorized by camera height. Each group is processed through a pretrained VGGT network~\cite{r12} to generate multi-scale 3D point clouds, which utilizes an end-to-end Transformer architecture to directly predict 3D attributes from multi-scale inputs:
\begin{equation}
    PC^{s1}, PC^{s2}, PC^{s3} = F_{VGGT}(I_d^{s1}, I_d^{s2}, I_d^{s3}),
\end{equation}
where ${s_1, s_2, s_3}$ correspond to different scales: low, medium, and high, respectively. The multi-scale point clouds capture complementary perspectives of the target: higher viewpoints emphasize global structural features, while lower viewpoints preserve finer local details.

To enhance point cloud representation, we utilize PointCLIP~\cite{r6} as a frozen feature extractor. Recognizing the fundamental differences between 3D point clouds and 2D images, PointCLIP projects the point clouds into multi-view depth maps to enable compatibility with CLIP's image encoder. These multi-view features are then fused with the original CLIP features through residual-connected linear adapter layers, forming an enriched global representation:
\begin{equation}
    f_d^{global}=\sigma(\text{Concat}(f_d^1\cdots f_d^M)W_3)W_4,
\end{equation}
\begin{equation}
    f_d^{si} = f_d^{global}+\sigma(f_d^{global}W_5),
\end{equation}
where $M$ represents the number of views, $W_3$, $W_4$, and $W_5$ denotes different weight matrices. The multi-scale vectors are aligned with $f_g$ and $f_s$ to maintain cross-view semantic consistency.

\subsection{Loss Function}

We propose a cross-view feature alignment module to bridge the street-satellite viewpoint gap using 3D representations. For street feature $f_g$, satellite feature $f_s$ and pointcloud feature $\{f_d^{si}\}_{i=1}^3$, we use InfoNCE loss $\mathcal{L}_{info}(i, j)=-\text{log}\frac{\text{exp}(\text{sim}(f_i, f_j)/\tau)}{\sum_{k=1}^K\text{exp}(\text{sim}(f_i, f_j^k)/\tau)}$ to constrain the similarity between the two of them. The cross-view contrast loss $\mathcal{L}_{cc}$ is fomulated as follows:
\begin{equation}
    \mathcal{L}_{cc}=\mathcal{L}_{info}(f_g,f_s)+\sum_{i=1}^{3}(\mathcal{L}_{info}(f_g,f_d^{si})+\mathcal{L}_{info}(f_s,f_d^{si})).
\end{equation}

Since intra-view iamges can also have mismatched viewpoints (e.g., front and side views of a house, etc.), and they have similar semantic representations. Therefore we design the self-supervised contrastive loss $\mathcal{L}_{sc}$ to constrain the intra-view images:
\begin{equation}
    \mathcal{L}_{sc}=\mathcal{L}_{info}(f_g,f_g)+\mathcal{L}_{info}(f_s,f_s)+\sum_{i=1}^3\mathcal{L}_{info}(f_d^{si},f_d^{si}).
\end{equation}

$\mathcal{L}_{sc}$ encourages instance-level separation by contrasting the positive pair against all other negatives in the batch. Even when the positive pair is identical, the optimization still acts on reducing similarities with negative samples, thereby enhancing intra-view discriminability.

Finally, the total loss function can be formulated as below:
\begin{equation}
\mathcal{L}_{total}=\mathcal{L}_{cc}+\lambda\mathcal{L}_{sc},
\end{equation}
where $\lambda$ is the weight factor to balance the contribution of $\mathcal{L}_{sc}$.

\setlength{\tabcolsep}{5.8pt}
\begin{table}[t]
\renewcommand\arraystretch{1.2}
\small
\begin{center}
\begin{tabular}{c|c|cccc}
\hline \hline
\textbf{Methods} & \textbf{Backbone} & \textbf{R$@$1} & \textbf{R$@$5} & \textbf{R$@$10} & \textbf{AP} \\
\hline
LPN \cite{r2} & ResNet50 & 0.81 & - & - & 2.52 \\
PLCD \cite{r3} & ResNet50 & 9.51 & 27.66 & 38.83 & 14.16 \\
Sample4Geo \cite{r21} & ConvNext-B & 0.89 & 3.61 & 5.43 & 1.87 \\
CVcities \cite{r4} & DINOv2-B & 15.74 & 33.58 & 44.47 & 20.24 \\ \hline
SkyLink* & DINOv2-B & 18.65 & 37.15 & 47.69 & 23.48 \\
SkyLink & DINOv2-L & \textbf{27.06} & \textbf{52.46} & \textbf{64.17} & \textbf{33.09} \\
\hline \hline
\end{tabular}
\end{center}
\caption{ Comparison with the other methods under Ground $\to$ Satellite setting. R@K (\%) is Recall@K, and AP (\%) is average precision. The best results are marked in bold.}
\label{table:Generic}
\vspace{-.25in}
\end{table}

\section{Experiment}
\subsection{Implementation Details}
All experiments are conducted on two NVIDIA RTX 4090 24G GPUs under PyTorch framework. We train on the University-1652 dataset, augmenting its original street-view with supplemental Google imagery, which incorporating an additional 2,463 street-view images provided by GREM. The total number of training epoch is 40. The framework employs DINOv2-L as the backbone to extract 4096-dimensional 2D features, optimized via stochastic gradient descent (momentum $0.9$) with cosine-annealed learning rates varing from $5 \times 10^{-4}$ to $1 \times 10^{-4}$ and $10\%$ warmup. For the weight factor $\lambda$, the value is set to 3.0. All input images are resized to $448 \times 448$ and augmented with random JPEG compression, color jitter, blur/sharpening filters, grid/coarse dropout, and normalization. Random rotation is further conducted for satellite-view imagery. During inference, we extract features using the frozen backbone and enhance robustness through test-time augmentation (TTA), including horizontal flipping and rotate90 augmentation.

\subsection{Comparison With Other Methods} \label{sec:localization}
To evaluate our method, we retrieve satellite images from street-view queries. As shown in Table~\ref{table:Generic}, on the University-1652 dataset, our approach achieves state-of-the-art results in ground-to-aerial retrieval, with R@1 of 27.06\%, R@5 of 52.46\%, R@10 of 64.17\%, and AP of 33.09\%, significantly surpassing previous methods. Compared to LPN and PLCD with ResNet50 backbones, our method improves R@1 by over 26\% and 17\%, respectively. Even with the smaller DINOv2-B model, it outperforms existing DINO-based methods like CVcities and Sample4Geo, validating its effectiveness in challenging ground-to-aerial matching tasks.



\setlength{\tabcolsep}{5.5pt}
\begin{table}
\small
\begin{center}
\begin{tabular}{cccc|cccc}
\hline \hline
PAFA & SSL & MSBM & GREM & R$@$1 & R$@$5 & R$@$10 & AP \\
\hline
\ding{51} & \ding{55} & \ding{55} & \ding{55} & 18.42 & 36.91 & 45.73 & 23.27 \\
\ding{51} & \ding{51} & \ding{55} & \ding{55} & 20.91 & 42.55 & 51.18 & 26.21 \\
\ding{51} & \ding{55} & \ding{51} & \ding{55} & 23.63 & 46.82 & 56.79 & 30.63 \\
\ding{51} & \ding{51} & \ding{51} & \ding{55} & 26.64 & 51.26 & 63.68 & 32.64 \\
\ding{51} & \ding{51} & \ding{51} & \ding{51} & \textbf{27.06} & \textbf{52.46} & \textbf{64.17} & \textbf{33.09} \\
\hline \hline
\end{tabular}
\end{center}
\caption{Ablation study of different components in our proposed framework. The best results are marked in bold.
}
\label{table:comp}
\vspace{-.2in}
\end{table}

\setlength{\tabcolsep}{12pt}
\begin{table}
\small
\begin{center}
\begin{tabular}{c|cccc}
\hline \hline
Methods & R$@$1 & R$@$5 & R$@$10 & AP \\
\hline
NetVLAD \cite{r7} & 25.66 & 50.88 & 63.11 & 31.58 \\
Gem Pool \cite{r8} & 25.78 & 51.04 & 63.37 & 32.40 \\
Conv-AP \cite{r9} & 26.38 & 51.28 & 63.49 & 32.69 \\
PAFA (Ours) & \textbf{27.06} & \textbf{52.46} & \textbf{64.17} & \textbf{33.09} \\
\hline \hline
\end{tabular}
\end{center}
\caption{Ablation study of different modules for feature aggeration part. The best results are marked in bold.
}
\label{table:agg}
\vspace{-.2in}
\end{table}

\setlength{\tabcolsep}{15pt}
\begin{table}
\small
\begin{center}
\begin{tabular}{c|cccc}
\hline\hline
$\lambda$ & R$@$1 & R$@$5 & R$@$10 & AP \\
\hline
1.0 & 24.50 & 50.46 & 61.74 & 30.08 \\
2.0 & 26.10 & 51.53 & 62.49 & 31.25 \\
3.0 & \textbf{27.06} & \textbf{52.46} & \textbf{64.17} & \textbf{33.09} \\
4.0 & 25.81 & 51.28 & 62.28 & 31.86 \\
5.0 & 25.27 & 50.90 & 61.81 & 31.59 \\
\hline\hline
\end{tabular}
\end{center}
\caption{The sensitivity analysis of the hyperparameter $\lambda$. The best results are marked in bold.
}
\label{table:loss}
\vspace{-.2in}
\end{table}


\subsection{Ablation Study} \label{sec:ablation}
The cross-view contrastive loss is widely applied in previous works \cite{r16}. Therefore, $\mathcal{L}_{cc}$ is used as the basic loss objective in the section. Ground $\rightarrow$ Satellite conducted on the test dataset is used to compare the corrsponding experiments.

\noindent\textbf{Effect of different components in the framework.} We leverage PAFA for feature aggregation in this section. As shown in Table~\ref{table:comp}, SSL improves metrics via contrastive supervision, while MSBM enhances cross-scale point cloud alignment for further gains. Combining SSL and MSBM achieves the best results with 27.06\% R@1 and 33.09\% AP. GREM further boosts all metrics, highlighting the value of diverse street perspectives. These components work synergistically to enhance cross-view retrieval accuracy.

\noindent\textbf{Effect of different modules for feature aggeration part.} We conduct ablation studies on the PAFA module in Table~\ref{table:agg}, comparing with three widely-used aggregation methods: NetVLAD, Gem Pool, and Conv-AP. Experimental results show that our PAFA module consistently outperforms all baselines across multiple metrics, which demonstrates the effectiveness of our PAFA strategy in capturing discriminative and robust global image representations.

\noindent\textbf{Senstivity of the hyperparameter $\lambda$.} We evaluate the impact of the self-supervised contrastive loss by varying the hyperparameter $\lambda$ from 1.0 to 5.0. As shown in Table~\ref{table:loss}, performance improves as $\lambda$ increases, peaking at $\lambda=3.0$, where R@1 and AP peak at 27.06\% and 33.09\%, respectively. However, further increasing $\lambda$ degrades performance, indicating that overemphasizing the contrastive term may hinder the learning of discriminative cross-view cues.

\subsection{Visualization.} \label{sec:vis}
\begin{figure}[t]
\begin{center}
     \includegraphics[width=0.95\linewidth]{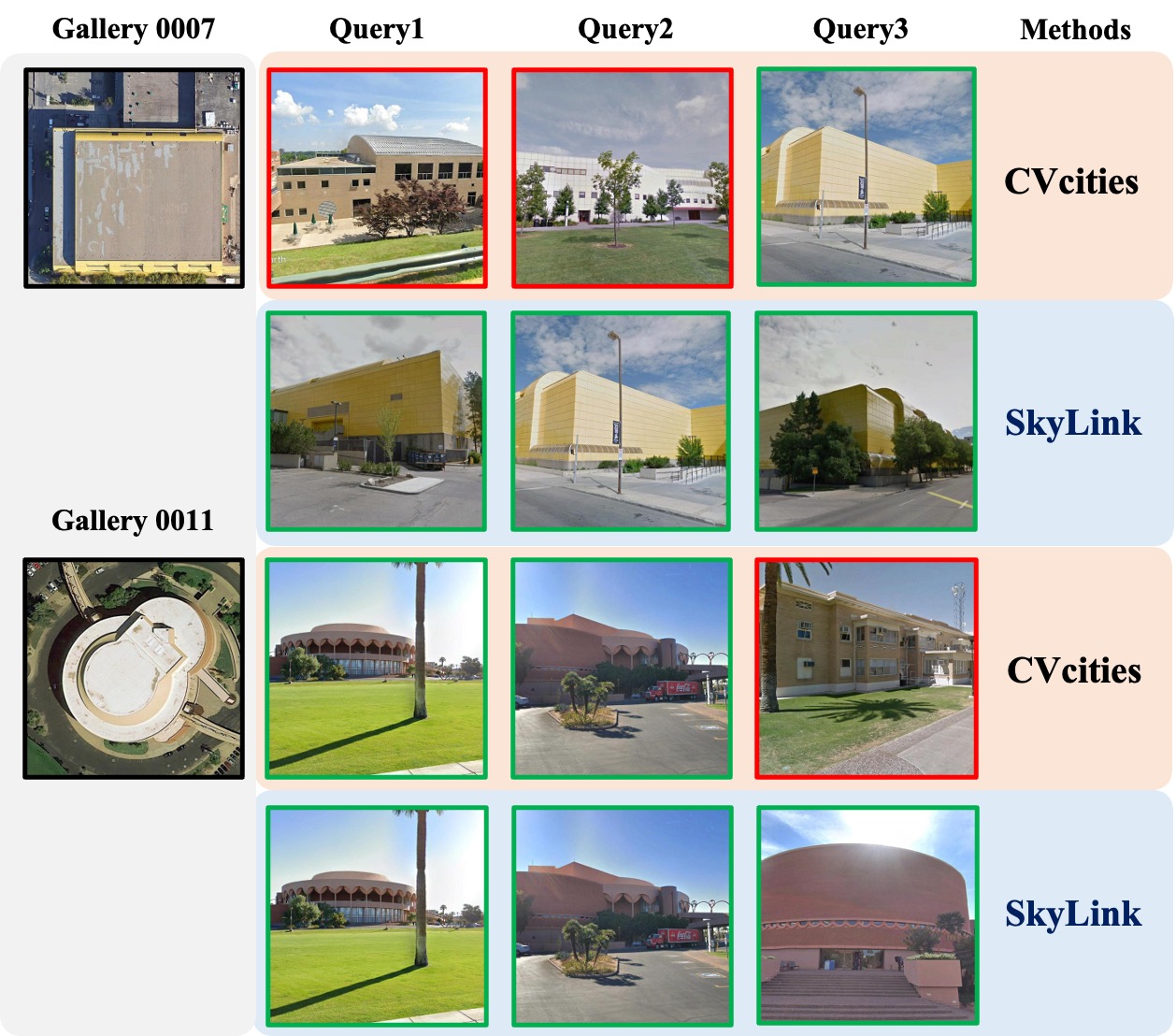}
\end{center}
      \caption{ Qualitative Street-to-satellite retrieval results between CVcities and our method. The green and red boxes indicate correct and error retrieval results, respectively.
      }\label{fig:retrieval}
\vspace{-.2in}
\end{figure}
For additional qualitative evaluation, we show retrieval results by our model on University-1652 test set. As illustrated in Figure~\ref{fig:retrieval}, the CVcities method often retrieves visually similar but incorrect matches, indicating its limited ability to capture fine-grained cross-view correspondences. In contrast, our method consistently retrieves the correct ground-view images that accurately correspond to the aerial queries, demonstrating stronger spatial understanding and robustness to viewpoint changes.

\section{Conclusion}
In this paper, we present a novel framework SkyLink for cross-view geo-localization by effectively matching street-view images with satellite and drone imagery. The GREM is introduced to enrich ground-level views through realistic data augmentation. Additionally, multi-view UAV images are exploited to construct 3D scenes, bridging the large perspective gap between ground and satellite. The proposed PAFA module further improves global feature representation. Extensive experiments demonstrate that our method achieves robust performance across diverse geo-localization, obtaining excellent performance on University-1652 in the UAVM2025 Challenge.

\bibliographystyle{ACM-Reference-Format}
\balance
\bibliography{sample-base}










\end{document}